\documentclass[sigconf]{acmart}
\AtBeginDocument{%
  \providecommand\BibTeX{{%
    \normalfont B\kern-0.5em{\scshape i\kern-0.25em b}\kern-0.8em\TeX}}}

\copyrightyear{2021}
\acmYear{2021}
\setcopyright{acmcopyright}\acmConference[CVMP '21]{European Conference on Visual Media Production}{December 6--7, 2021}{London, United Kingdom}
\acmBooktitle{European Conference on Visual Media Production (CVMP '21), December 6--7, 2021, London, United Kingdom}
\acmPrice{15.00}
\acmDOI{10.1145/3485441.3485650}
\acmISBN{978-1-4503-9094-1/21/12}

\usepackage{mathtools}
\usepackage{graphicx}
\usepackage{xspace}
\DeclarePairedDelimiter\ceil{\lceil}{\rceil}
\begin{document}

\title{VPN: Video Provenance Network for Robust Content Attribution}

\author{Alexander Black}
\affiliation{%
  \institution{CVSSP, University of Surrey}
}
\email{alex.black@surrey.ac.uk}

\author{Tu Bui}
\affiliation{%
  \institution{CVSSP, University of Surrey}
}
\email{t.v.bui@surrey.ac.uk}

\author{Simon Jenni}
\affiliation{%
  \institution{Adobe Research}
}
\email{jenni@adobe.com}

\author{Vishy Swaminathan}
\affiliation{%
  \institution{Adobe Research}
}
\email{vishy@adobe.com}

\author{John Collomosse}
\affiliation{%
  \institution{University of Surrey | Adobe Research}
}
\email{collomos@adobe.com}

\renewcommand{\shortauthors}{Black, et al.}

\begin{abstract}
  We present VPN - a content attribution method for recovering provenance information from videos shared online.  Platforms, and users, often transform video into different quality, codecs, sizes, shapes, etc. or slightly edit its content such as adding text or emoji, as they are redistributed online.  We learn a robust search embedding for matching such video, invariant to these transformations, using full-length or truncated video queries.  Once matched against a trusted database of video clips, associated information on the provenance of the clip is presented to the user. 
  We use an inverted index to match temporal chunks of video using late-fusion to combine both visual and audio features.  In both cases, features are extracted via a deep neural network trained using contrastive learning on a dataset of original and augmented video clips.  We demonstrate high accuracy recall over a corpus of 100,000 videos. 

\end{abstract}



 \begin{teaserfigure}
 \centering
   \includegraphics[width=\linewidth]{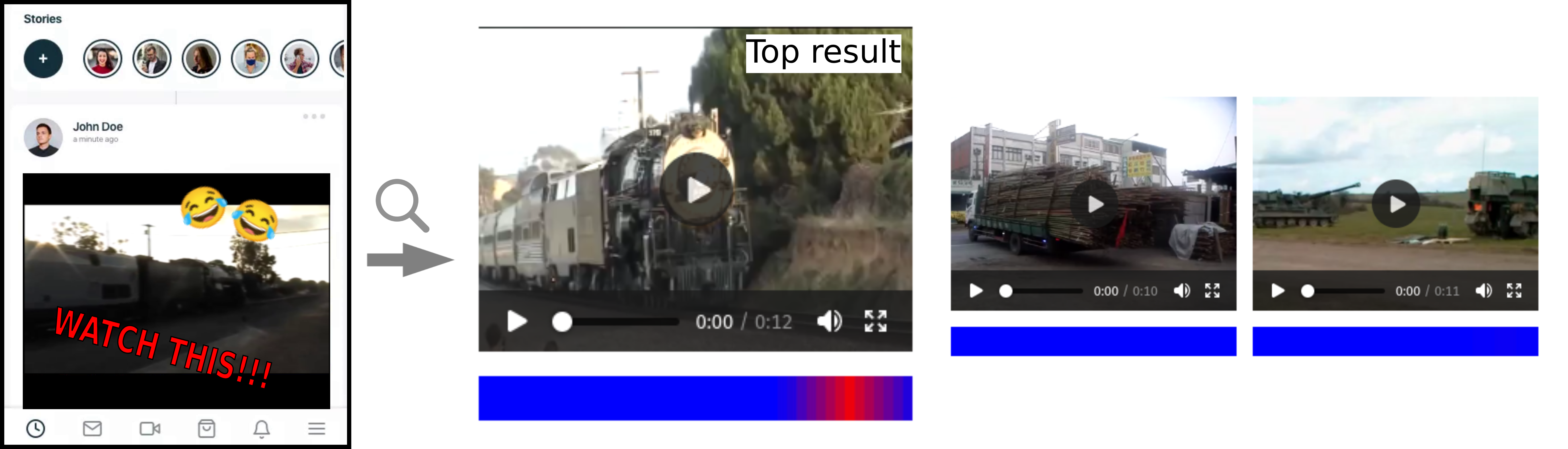}
   \caption{Videos circulating on social media are matched to a trusted database containing media provenance information, enabling the consumer to make informed trust decisions on the content. Matching is robust to audio-visual transformations commonly introduced during redistribution e.g. quality, resolution, or format change as well as partial clips (truncation).}
   \label{fig:teaser}
 \end{teaserfigure}

\maketitle
\makeatletter
\DeclareRobustCommand\onedot{\futurelet\@let@token\@onedot}
\def\@onedot{\ifx\@let@token.\else.\null\fi\xspace}

\def\eg{\emph{e.g}\onedot} \def\Eg{\emph{E.g}\onedot}
\def\ie{\emph{i.e}\onedot} \def\Ie{\emph{I.e}\onedot}
\def\cf{\emph{c.f}\onedot} \def\Cf{\emph{C.f}\onedot}
\def\etc{\emph{etc}\onedot} \def\vs{\emph{vs}\onedot}
\def\wrt{w.r.t\onedot} \def\dof{d.o.f\onedot}
\def\etal{\emph{et al}\onedot}
\makeatother

\section{Introduction}
Video is a powerful medium for storytelling. Yet, the ease with which digital video may be synthesized, manipulated and shared (\eg via social media) presents a growing societal threat via the amplification of misinformation and fake news \cite{ticks}.

Emerging solutions counter this threat in two main ways: 1) Detecting synthetic or manipulated video \cite{kaggledf}. Detection methods are caught in a perpetual race against generative AI tools that seek to evade detection.  Moreover,  digital manipulation is often for editorial rather than deceptive purposes; 2) Determining the `{\em provenance}' of the video asset \cite{Black_2021_CVPR,nguyen2021}.  Provenance techniques do not seek to adjudicate on content validity but instead trace the origins of content, and what was done to it. Rather than automatically deciding on the trustworthiness of the content, the goal is to place the consumer in a more informed position to make a trust decision.

This paper falls into the second camp,  and complements emerging technical standards that embed  cryptographically secured provenance information with the metadata of the asset \cite{c2pa,cai,origin}.  For example, the emerging specification from cross-industry body the `Coalition for Content Provenance and Authenticity' (C2PA) writes provenance information into a `manifest' transported within the asset metadata \cite{c2pa}. 
Such approaches are vulnerable to removal of metadata, which is common on social media platforms through which misinformation is often spread.  For example, video uploaded to any major social media platform today would be stripped of such manifests.  Furthermore, alternative manifests may be substituted describing a fake provenance trail or `back story', so attributing a video out of context to tell a different story. Content misattribution may also deprive creators of credit for their work, enabling intellectual property theft.


This paper contributes a method for robustly matching video assets circulating {\em without provenance metadata}, to an authoritative copy of that asset with such metadata (such as a C2PA manifest), held within a trusted database.   Videos often undergo various transformations during online distribution; changes in format, resolution, size, padding, effect enhancement \etc that render cryptographic hashes operating on the binary stream, such as SHA-256, unsuitable as means for matching the video content. Rather, matching must be made robust to these transformations by considering features extracted from the content of the video clip.   It is also desirable to match fragments of video (\ie partial or truncated videos) to determine not only the complete source video but also the time offset at which that fragment exist.  

We present a system for matching partial video queries robust to such transformations.  We build an inverted index of robust audio-visual features trained using contrastive learning and a rich set of augmentations representative of transformations typically applied to video `in the wild' during online content distribution.  We leverage recent work exploring the visual comparison of image matches under similar transformations, and show this technique may be applied to visually compare a query and its matched results to identify areas of content manipulation.  We demonstrate our matching technique within the context of a  system for tracing the provenance of video assets, using a corpus of 100,000 videos from the VGGSound dataset \cite{vggsound}.

\section{Related work}

Video content authenticity has been explored primarily from the digital forensics perspective of detecting generative (or `deep fake’) content, or identifying if and where video has been digitally tampered. The Deep Fake Detection Challenge (DFDC) \cite{kaggledf}  explored detection of facial manipulation in video.  Networks may be trained to  characterize departures from natural image statistics \cite{zhang2020}, or artifacts introduced can be recognized to localize manipulated regions  \cite{zhang2019}, in some cases even identifying which GAN synthesized an image \cite{Yu2021}.  Detection of video manipulation similarly exploits temporal anomalies \cite{mantranet2019cvpr} or GAN limitations such as lack of blinking \cite{blink}.  Forensic models are in a perpetual `arms-race' with the rapid innovation of generative methods seeking to evade detection.

Attribution methods trace the provenance of media – where it came from, and what has been done to it.  A robust identifier, such as a {\em watermark} \cite{hameed2006,devi2009,profrock2006,baba2009}, may be embedded in media or a perceptual hash or {\em fingerprint} is computed from the media content itself \cite{iscc,khelifi2017,dsh2016cvpr,hashnet2017iccv,Moreira2018,Zhang2020manip,Bharati2021}.  The identifier serves as a key to lookup provenance information in a provenance database operated by a trusted source \cite{origin}, or in a decentralized trusted database such as a blockchain \cite{archangel,bui2020archangel}.

In the audio domain, music information retrieval (MIR) has been effective in matching audio or audio fragments at scale \cite{mir,mirtutorial}. Classical approaches analyze frequency distributions, e.g. short-time Fourier Transform (STFT) \cite{Grill2015},  Mel-spectrogram variants such as MFCC \cite{mfcc}, or Constant-Q transform \cite{Humphrey2012}, to identify music fragments.  More recently, convolutional neural networks (CNNs) have been applied to classify spectrogram representations of audio for instrument recognition \cite{Han2017}, on-set detection \cite{Schluter2013}, music tagging at scale \cite {Dieleman2014,Lee2017,millionsong} and genre recognition \cite{Lee2009,gtzan}.  Music similarity has been explored using feature embeddings regressed via CNN \cite{Lu2017}.  Recurrent neural networks (RNN) have also been applied to integrate such features for tagging \cite{Choi2014}, transcription \cite{Rigaud2016,Sigtia2015} and hit song prediction \cite{Yang2017}.

In the image domain, visual similarity algorithms focus upon compact, robust perceptual hashing \cite{hashsurvey}.  Early approaches compute content fingerprints from coefficients in the spectral domain  \cite{phash,iscc,jacobs1995}. CNNs are now the dominant method for image similarity.  Deep Hashing Networks (DHNs) \cite{dhn2016aaai} extended an ImageNet-trained AlexNet \cite{alexnet} feature encoder \cite{hadsell} with a quantization loss, to obtain hashes that retained semantic discrimination.  Deep Supervised Hashing (DSH) \cite{dsh2016cvpr} and HashNet \cite{hashnet2017iccv} also train CNNs to learn visual hashes, using a siamese network and ranking loss; such losses are better suited to retrieval than classification tasks,  and are used extensively in visual search \cite{gordo2016deep}. DSDH~\cite{dsdh2017nips} learns metric ranking and classification directly from the hash code using a discrete cyclic coordinate descend technique. DPSH~\cite{dpsh2016cjai} proposes a binary pairwise loss to estimate image pair similarity. DFH~\cite{dfh2019bmvc} and GreedyHash~\cite{greedy2018nips} use different customized layers to address the ill-posed gradient of the sign function. ADSH~\cite{adsh2018aaai} and CSQ~\cite{csq2020cvpr} treat hashing as a retrieval/attribution optimization problem, but require semantic annotation unavailable in the content attribution problem domain.

Visual similarity for attribution and content provenance was explicitly addressed by Nguyen \etal \cite{nguyen2021} who learn a compact binary hash robust to non-editorial transformation, but sensitive to digital manipulation in the image.  The goal was to match robustly but intentionally fail to match manipulated content.  Conversely, Black \etal \cite{Black_2021_CVPR} learn a fingerprint robust to both non-editorial and editorial change, opting instead to highlight regions of likely manipulation once matched.  In both cases provenance information for the asset is surfaced after the match.  Our work explores a similar matching strategy for video, and explores change visualization via \cite{Black_2021_CVPR} over  frames.  Our work combines robust matching of both audio and visual features, exploring appropriate fusion strategies and indexing to support partial asset (video fragment) matching.  We also train our features under data augmentations recently proposed for the task of image similarity for content attribution \cite{fbimagesimilarity}.

\section{Methodology}
\label{sec:method}

\begin{figure*}[t!]
    \centering
    \includegraphics[width=1.0\linewidth]{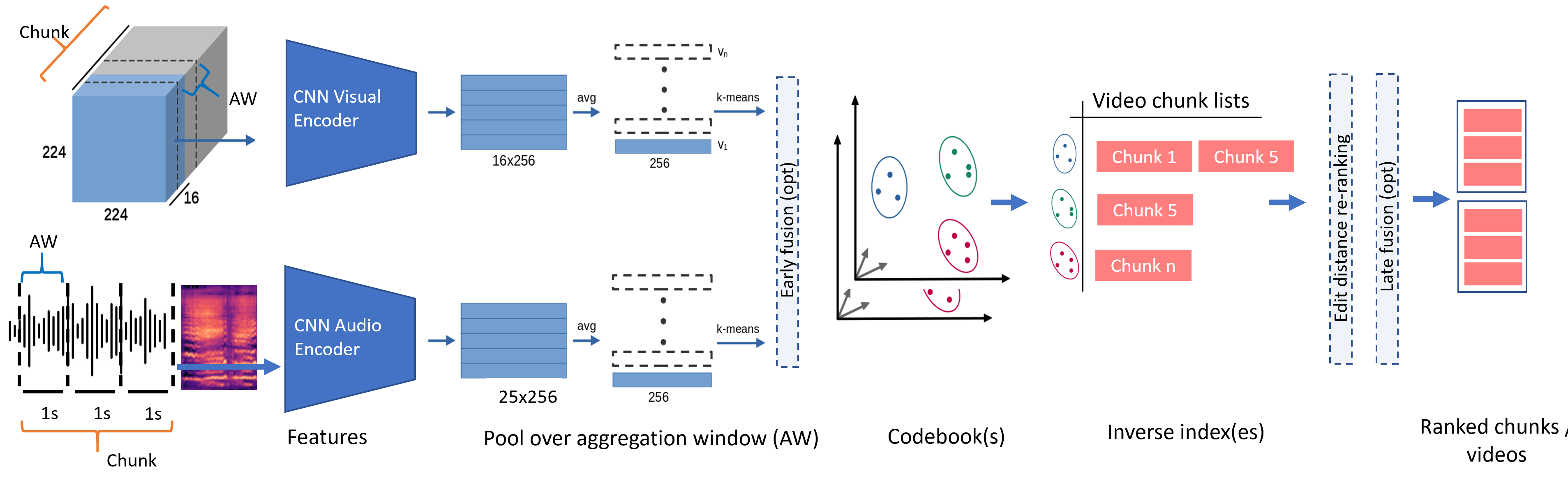}
    \caption{Architecture of the proposed system. Video is regularly sub-divided into equal-length chunks, in which a set of visual and audio descriptors is computed for each chunk. These independently computed features are regularly sampled from short temporal aggregation windows (AW) within the chunk. The visual and audio features are quantised to create a dictionary upon which an inverted index is built to index videos at chunk-level. We explore 3 fusion strategies for video and audio retrieval, that differ in which stage the audio and video features are combined. A re-ranking on edit distance yields the ranked list of video chunks returned to the user.}
    \label{fig:vis_arch}
\end{figure*}
To learn a robust video fingerprinting model, we propose a self-supervised network capable of encoding both visual and audio streams in a video. We leverage contrastive learning and a rich set of data augmentations for videos to train our model. To enable partial video matching, we follow a `divide and conquer' approach where video is split into chunks and each chunk is indexed and search-able within an inverted index. Our retrieval pipeline is shown in Figure~\ref{fig:vis_arch}. We describe our encoding model in sec.~\ref{sec:features}, our chunking, indexing method and ranking strategy in sec.~\ref{sec:index}, and our audio-visual fusion method in sec.~\ref{sec:fusion}.


\subsection{Learning robust visual-audio features}
\label{sec:features}
A video circulated on the Internet may undergo certain transformations that affect either the visual or audio stream or both. For example, the visual stream may be subjected to quality reduction (noise, blur, pixelization,...) during reformatting, change in aspect ratio/geometry (padding, resize, rotation, flip,...), visual enhancement (brightness, color adjustment,...) or editorial change (text/emoji overlay, photoshop,...). Similarly, the audio stream could also be altered (compression, background noise, trimming, effect enhancement, ...). We treat such transformations as {\em perturbations} to the original video and propose to learn a frame-level visual model and an audio model to encode the respective video streams robust to the defined perturbations. 

\subsubsection{Learning frame-level visual features}
\label{sec:features_visual}
We train a CNN model $f^v(.)$ to project a video frame to a compact embedding space. We employ the ResNet50 architecture \cite{he2016deep}, replacing the final classifier layer with a 256-D fully connected (fc) layer that serves as the embedding. Our model is trained with loss:
\begin{align}
    &\mathcal{L}(z) = -\log{\frac{e^{d(z,\bar{z}_+)/\tau}}{e^{d(z,\bar{z}_+)/\tau} + \sum_{z_-}{e^{d(z,z_-)/\tau}}}} \label{eq:simclr}\\
    \textrm{where } &d(u,v) = \frac{g(u) \cdot g(v)}{\left|g(u)\right|\left|g(v)\right|}
\end{align}
where $z$ is the embedding of a video frame $v$: $z=f^v(v) \in \mathbb{R}^{256}$; $\bar{z}_+$ is the average embedding of all transformations of $v$ in the mini-batch; $z_-$ denotes other frame instances; $g(.)$ is a set of two MLP layers separated by ReLU that acts as a buffer between the embedding and the loss function; $d(u,v)$ measures the cosine similarity between the intermediate embeddings $g(u)$ and $g(v)$; $\tau$ is the contrastive temperature ($\tau=0.1$ in our experiments). $\mathcal{L}(.)$ aims to bring the embeddings of all transformations of an image (frame) together, while pushing away other image instances. $\mathcal{L}(.)$ resembles NTXent loss \cite{chen2020simple}, however eq.~\ref{eq:simclr} accepts multiple positives in a batch instead of just a single pair of image augmentations.

We initialize our model with weights from DeepComp \cite{Black_2021_CVPR}. During training, we randomly sample frames from the training videos to construct a batch. For each frame image we create $p$ augmentations to serve as positive samples (to compute $\bar{z}_+$), while the rest in the batch acts as negatives (the denominator term in eq.~\ref{eq:simclr}). We empirically set $p=3$ for optimal performance (improves by 1\% as compared with standard NTXentLoss). We observe that larger $p$ causes a drop in performance, probably because the number of unique images in the batch must be reduced accordingly in order to fit a GPU.

\begin{figure}[t!]
    \centering
    \includegraphics[width=\linewidth]{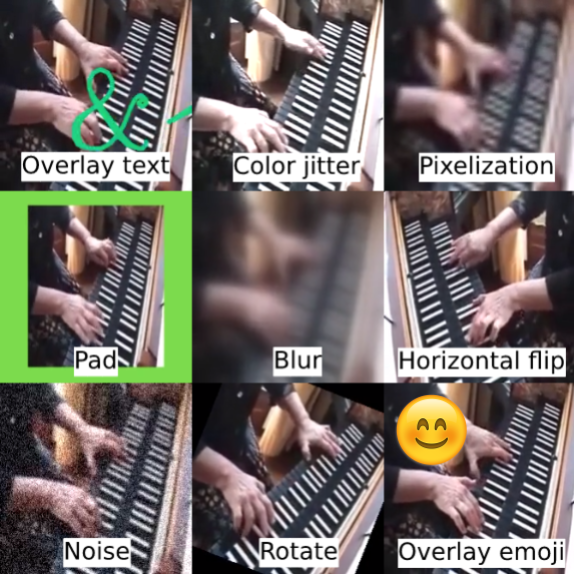}
    \caption{Visual augmentation: examples of random benign augmentations applied to a video frame during contrastive training of the visual feature encoder.}
    \label{fig:vis_aug}
\end{figure}

We use an exhaustive list of frame-level augmentations for our training, including random Noise (variance 0.01), Blur (radius $[0, 10]$), Horizontal Flip, Pixelization (ratio $[0.1, 1.0]$), Rotation ($[-30, +30]$ degrees), random Emoji Overlay (opacity $[80, 100]$\%, size $[10, 30]$\%, random position), Text Overlay (text length $[5, 10]$, size $[10, 20$]\%, random characters, typeface and position), Color Jitter (brightness, contrast and saturation $[0.6, 1.4]$), Padding ($[0, 25]$\% dimension, random color). These are on top of 16 Imagenet-C \cite{imagenetc} transformations seen by the pretrained model \cite{Black_2021_CVPR} which our model initialises from. Since our model operates on individual video frames, all transformations are applied at frame level, \ie, the temporal coherence between frames are ignored during data augmentation. However, at test time our query videos are transformed at video level to reflect the video editing and distribution practice. Fig.~\ref{fig:vis_aug} illustrates several benign transformations applied to an example video frame. 

\begin{figure*}[t!]
    \centering
    \includegraphics[width=1.0\linewidth]{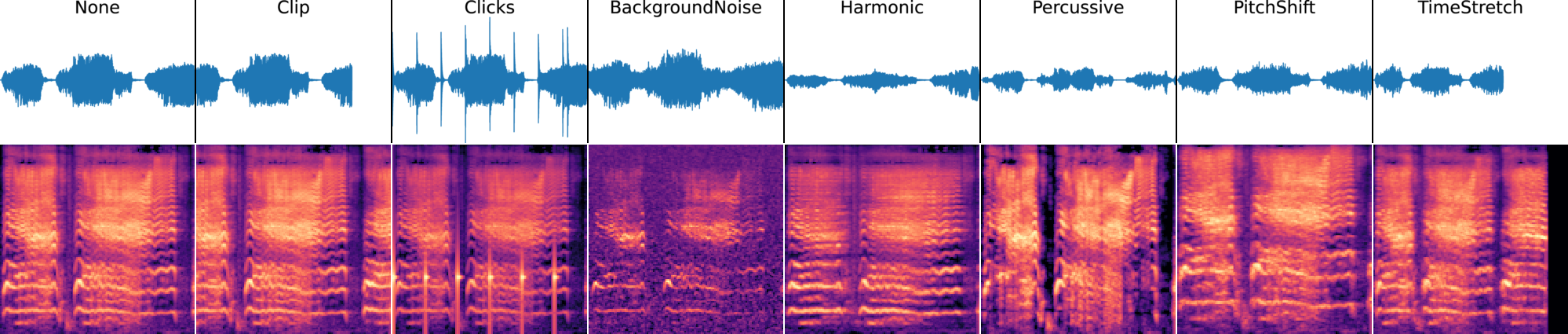}
    \caption{Audio augmentations: Magnitude (top) and mel-spectrogram (bottom) of an audio segment and random benign transformations used during training the audio feature encoder. }
    \label{fig:aud_aug}
\end{figure*}

\subsubsection{Learning audio features}
\label{sec:features_audio}
We split the audio signal of a video into overlapping 1-second chunks and encode it via log mel-spectrogram. We then visualize the log mel-spectrogram as a 2D RGB image and treat it as input to our audio model $f^a(.)$. The same model architecture and loss as in sec.~\ref{sec:features_visual} are employed for our training, but using a different set of data augmentation methods to learn robust audio features. In general, the benign audio transformations can be categorized in two groups - those that lengthen or shorten the audio signal and those that add, remove or alter audio components. The former includes audio Clipping ($[0, 20]$\% audio length) and Time Stretching (slow down 0.5x - speed up 1.5x). The latter includes Adding Clicks (random click rate 0.5sec - full length), Adding Background Noise (SNR 5db), Filtering Harmonics (margin $[1.0, 1.5]$), Filtering Percussive (margin $[1.0, 1.5]$) and Pitch Shifting (semitone $[-5, 5]$). These transformations are commonly encountered during audio redistribution and editing practice. Fig.~\ref{fig:aud_aug} shows the effects of these transformations on an example audio segment.


\subsection{Chunking and Indexing}
\label{sec:index}

To index a variable length video X, we propose to split X into fixed-length chunks $X=\{x_i|i=1,2,...,N\} \, s.t. \, len(x_i)=len(x_j)=l \quad \forall i,j \in [1,N]$ where $len(.)$ is the length function (in seconds), constant $l$ is the chunk length ($l$=10sec) and $N$ is number of chunks (the last chunk is padded if necessary). We split videos into chunks in a sliding window fashion with chunk stride $s_c \leq l$, thus $N=\ceil{len(X)/s_c}$. We use chunks as an atomic unit in our approach where a `bag of features' is computed for each chunk, also indexing and search are performed at chunk level. We use $s_c=l/2$ in our experiments.

\subsubsection{Computing chunk features}
Given a video chunk $x_i=\{x_i^v, x_i^a\}$ containing a visual stream $x_i^v$ and an audio stream $x_i^a$ of the same length $l$, we feed the two streams into the respective visual (sec.\ref{sec:features_visual}) and audio (sec.\ref{sec:features_audio}) models, obtaining a set of descriptors for both streams. For the visual stream, we sample $x_i^v$ at 16fps with stride $s_f$ ($s_f=0.5$ second or 8 frames), extract and average CNN features on every 16-frame aggregation window (AW) to get one visual descriptor per second. 
\begin{equation}
    z_i^v = \mathcal{F}^v(x_i^v) \in \mathbb{R}^{n\times 256},\quad n=\ceil{\frac{l}{s_f}}
\end{equation}
where $\mathcal{F}^v$ is our aggregation function, at sampling point j $\mathcal{F}^v(x_{i,j}^v)=\frac{1}{16}\sum_{t=0}^{15}{f^v(x^v_{i,j+t})}$; $n$ is number of visual descriptors per chunk. 

For the audio stream, since our audio model has input size of 1 second audio length, we sample $x_i^a$ at 1 second interval with the same stride $s_f$ as in the visual model. 
\begin{equation}
    z_i^a = f^a(x_i^a) \in \mathbb{R}^{n\times 256}
\end{equation}
This makes our audio extraction in sync with the visual extraction process (both have an aggregation window of 1 second), resulting in the same number of audio and visual descriptors per video chunk $x_i$. This setting also offers a level of flexibility to our audio-fusion strategy (sec.\ref{sec:fusion}).




\subsubsection{Inverted index}
\label{sec:invert}
Similar to text search systems, we construct an inverted index that supports video retrieval at chunk-level. We sample 1M random descriptors (audio, visual or fusion, more details in sec.~\ref{sec:fusion}) and build a dictionary with codebook size K using KMeans \cite{kmeans}. Given a database video, we break it into chunks where each chunk is represented as a bag of codewords. The K codewords are used as entries to our inverted index, listing all chunks in the database (a mapping between a chunk and video ID is also stored). 

The relevance of a query chunk $q = \{q_1, q_2, ..., q_n\}$ to a database chunk $x_i$ is determined by a \textit{relevance score} $\mathcal{R}$, defined as:

\begin{equation}
    \mathcal{R}(q, x_i) = \sum^{n}_{t=1}{\mathit{tf}(q_t, x_i) \times \mathit{idf(q_t)}}
    \label{eq:rel}
\end{equation}

where $\mathit{tf}(q_t, x_i)$ is the \textit{term frequency} that denotes the number of times codeword $q_t$ appears in the video chunk $x_i$ and $\mathit{idf}(q_t)$ is the \textit{inverse document frequency}, which measures how common is $q_t$ across all of the chunks in the dataset.

Finally, the relevance of a query video Q to a database video X is defined as:
\begin{equation}
    \mathcal{R}(Q, X) = \sum^{Q}_{q}{\sum_{x_i}^X{\mathcal{R}(q,x_i)}}
    \label{eq:rel_vid}
\end{equation}

\begin{figure*}[t]
    \centering
    \includegraphics[width=1.0\linewidth]{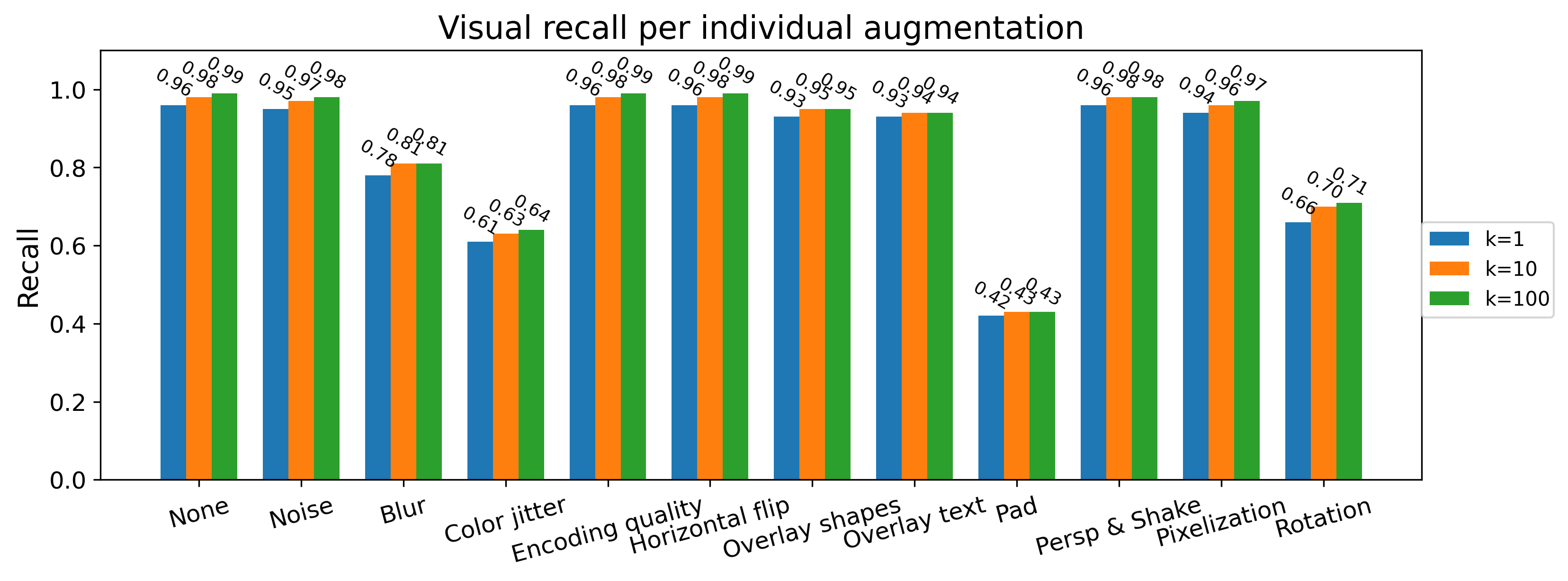}
    \caption{Visual retrieval performance breakdown per individual augmentations applied to the query, showing recall at $k=1, 10, 100$.}
    \label{fig:vis_iaug}
\end{figure*}

\begin{figure}
    \centering
    \includegraphics[width=1.0\linewidth]{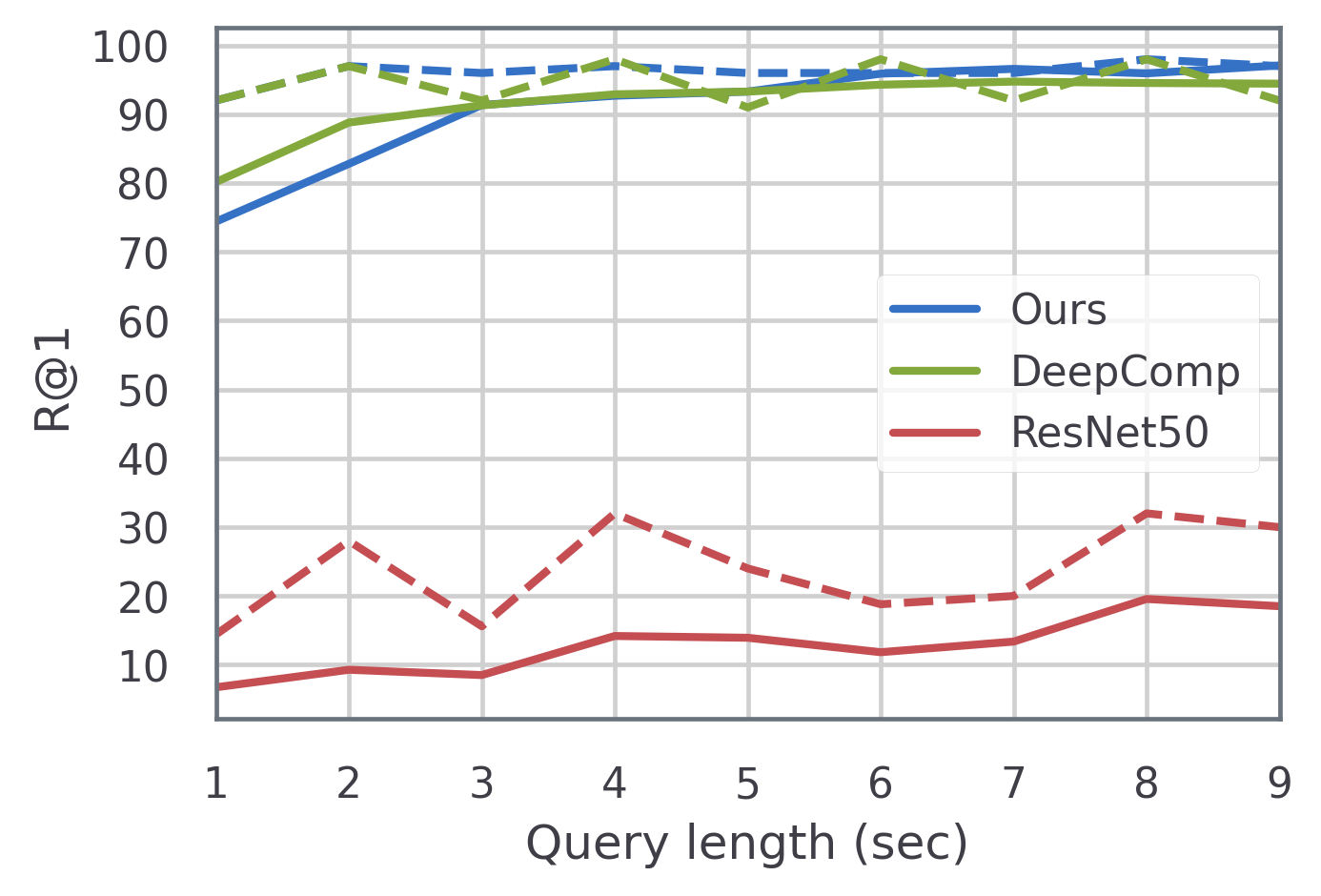}
    \caption{Visual retrieval performance showing R@1 and R@100 (dashed) in \% vs query segment length. Evaluated on VGGSound dataset.}
    \label{fig:vis_qlen}
\end{figure}

\begin{table}
    \centering
    \begin{tabular}{l|ccc}
        Method & R@1 & R@10 & R@100 \\ \hline
        Ours & 0.973 & 0.980 & 0.989 \\
        DeepComp \cite{Black_2021_CVPR} & 0.922 & 0.945 & 0.963 \\
        ResNet50 \cite{he2016deep} & 0.238 & 0.314 & 0.411 \\
    \end{tabular}
    \caption{Visual performance}
    \label{tab:vis_main}
\end{table}

\subsubsection{Re-ranking}
\label{sec:lev}
Inverted index does not take into account the order of descriptors within a chunk, and chunk order within a video. We therefore employ TF-IDF \cite{jones1972statistical} to retrieve top-k candidate videos (k=200) before performing an additional re-ranking stage based on Levenshtein distance \cite{levenshtein1966binary} which quantifies the similarity between two sequences by counting the number of {\em edits} (insertions, deletions or substitutions) required to turn one sequence into the other. Sorting by edit distance promotes to the top videos in which visual words appear in the order that matches the query the closest.

\subsubsection{Partial video matching and localization}
An advantage of our chunking and inverted index method is that it enables retrieval even if the query is only a segment of a database video. As a by product, our method also supports localization of a video segment by searching for the closest chunk in the database, or even the closest codeword within a chunk for more fine-grained localization. 

\begin{figure*}[t]
    \centering
    \includegraphics[width=1.0\linewidth]{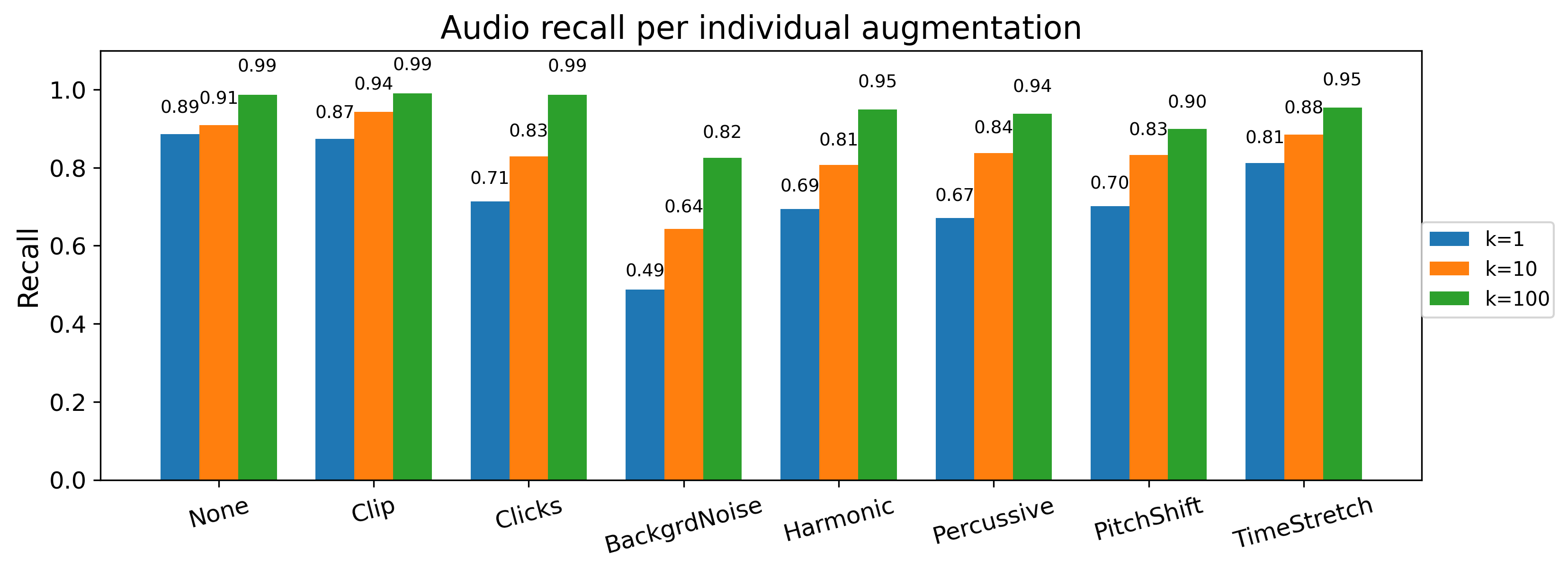}
    \caption{Audio performance on individual augmentations.}
    \label{fig:aud_iaug}
\end{figure*}

\begin{figure}
    \centering
    \includegraphics[width=1.0\linewidth]{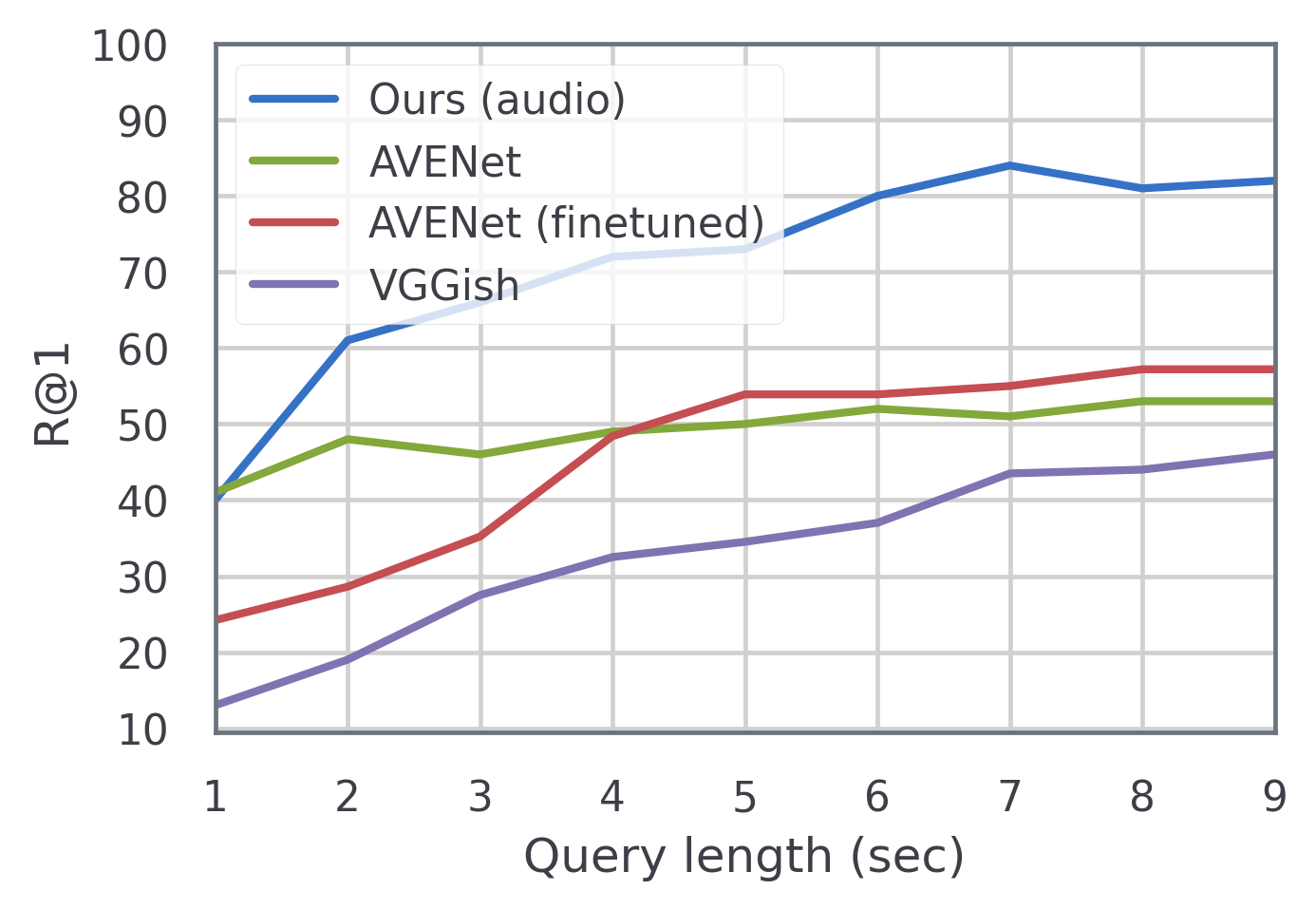}
    \caption{Audio performance on query segment length.}
    \label{fig:aud_qlen}
\end{figure}

\begin{table}
    \centering
    \begin{tabular}{l|ccc}
        Method & R@1 & R@10 & R@100 \\ \hline
        Ours & 0.826 & 0.899 & 0.969 \\
        AVENet (finetuned) & 0.572 & 0.688 & 0.728 \\
        AVENet \cite{vggsound} & 0.530 & 0.626 & 0.671 \\
        VGGish \cite{vggish} & 0.492 & 0.579 & 0.610
    \end{tabular}
    \caption{Audio performance}
    \label{tab:aud_main}
\end{table}

\subsection{Audio-visual fusion}
\label{sec:fusion}
We explore 3 fusion approaches for audio and visual streams to create an unified video retrieval system.

\noindent \textbf{Early fusion} - a single codebook is constructed for $z=[z_i^v, a_i^v] \in \mathbb{R}^{512}$, which is a concatenation of the visual and audio descriptor of a single aggregation window. 

\noindent \textbf{Late fusion} - we build separate codebooks and inverted indices for audio and visual domains, but compute the relevance score in eq.~\ref{eq:rel_vid} and re-ranking jointly.

\noindent \textbf{Learned fusion} - we learn an unified audio-visual embedding for a video AW. Since the visual model $f^v(.)$ operates at frame-level, we average the embeddings of the frames within an AW to represent the visual feature of that AW, before concatenating with the audio feature and projecting to the unified embedding:
\begin{equation}
    z = E_p([\frac{1}{\left | AW \right|}\sum_{x^v}^{AW}{f^v(x^v)}, f^a(x^a)]) \in \mathbb{R}^{256}
\end{equation}
where $E_p$ is a fc layer for dimensional reduction; $[,]$ is a concatenation; $\left | AW \right|$ is number of frames in an AW (to make the model small, we sample video at 4fps, thus $\left | AW \right|=4$). To train our model, we first train the audio and visual models separately, then use their weights to initialize our joint model training.



\section{Results and discussion}
\subsection{Datasets and evaluation metrics}
We use the following 2 datasets for training and evaluation:

\noindent \textbf{VGGSound} \cite{vggsound} is a large audio dataset with 160k train and 14k test videos. All videos are of 10 seconds length of 310 audio classes, originally used for audio classification. We train all of our models using the public VGGSound train set without using class annotations. To create a test set for retrieval, we first construct a 100k video database from the 14k test videos plus training videos to serve as distractors. Next, we sample 1k random videos from the test set and apply random transformations on both audio and video streams in order to make the query set. For experiments with query length, we randomly clip the queries keeping a minimum 10\% length (1 second). We use this dataset for the majority of our experiments.

\noindent \textbf{DFDC-Preview} \cite{dfdc_preview} consists of 5k original and manipulated videos. The dataset has a public train/test split and all videos have 15 seconds length. We employ DFDC-Preview to evaluate our model generalization to new domains (people and human voice) and unseen transformations (face swap manipulations). We do not retrain our model on DFDC-Preview, instead we evaluate our VGGSound-trained model directly on it. We use all 1131 real videos in DFDC-Preview together with distracting videos from VGGSound to make a 100k database. We create two query sets - a {\em real} set containing 276 real test videos that are pre-applied transformations with reduction in fps, resolution and encoding quality; and a {\em fake} set containing 501 face-swap manipulated videos. We wish to attribute the queries to the original videos in the database regardless of it being real or fake. 

Since there is only 1 relevant video in the database for each query, we use recall at k ($R@k$) to evaluate the retrieval performance of our approach. R@k represents the ratio of queries for which the correct video was retrieved within top $k$ returned results.

\subsection{Experiment settings}
We train our models using the Pytorch library, with Adam optimizer, initial learning rate 0.0005 with step decay and early stopping schedule. We use a batch size of 32 unique visual or audio samples where each sample is randomly augmented 3 times (totally 96, sec.~\ref{sec:features_visual}-\ref{sec:features_audio}). For chunking and indexing, since both VGGSound and DFDC-Preview videos are short and have equal length, we set the chunk size the same as the video length (note that our settings in sec.~\ref{sec:index} allow indexing and querying of arbitrary-length videos). For inverted index, we set the number of entries equal 100k, which is also the codebook size for both visual and audio modalities.

\begin{figure*}[t]
    \centering
    \includegraphics[width=1.0\linewidth]{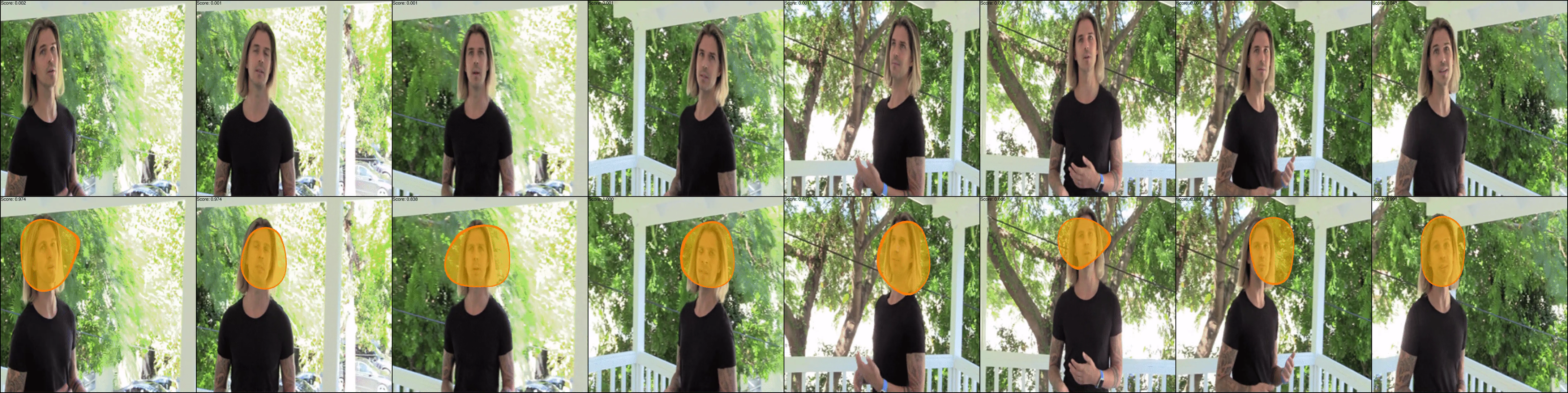}
    \caption{Using DFDC fake video (bottom) as query, we are able to retrieve the original (top) and use ICN\cite{Black_2021_CVPR} to visualize the manipulated areas.}
    \label{fig:dfdc_comp}
\end{figure*}

\begin{table}
    \centering
    \begin{tabular}{l|ccc}
        Method & R@1 & R@10 & R@100 \\ \hline
        Late fusion & 0.982 & 0.991 & 0.996 \\
        Learned fusion & 0.941 & 0.949 & 0.956 \\
        Early fusion & 0.789 & 0.842 & 0.913 
    \end{tabular}
    \caption{Fusion performance}
    \label{tab:fusion}
\end{table}

\begin{figure}
    \centering
    \includegraphics[width=1.0\linewidth]{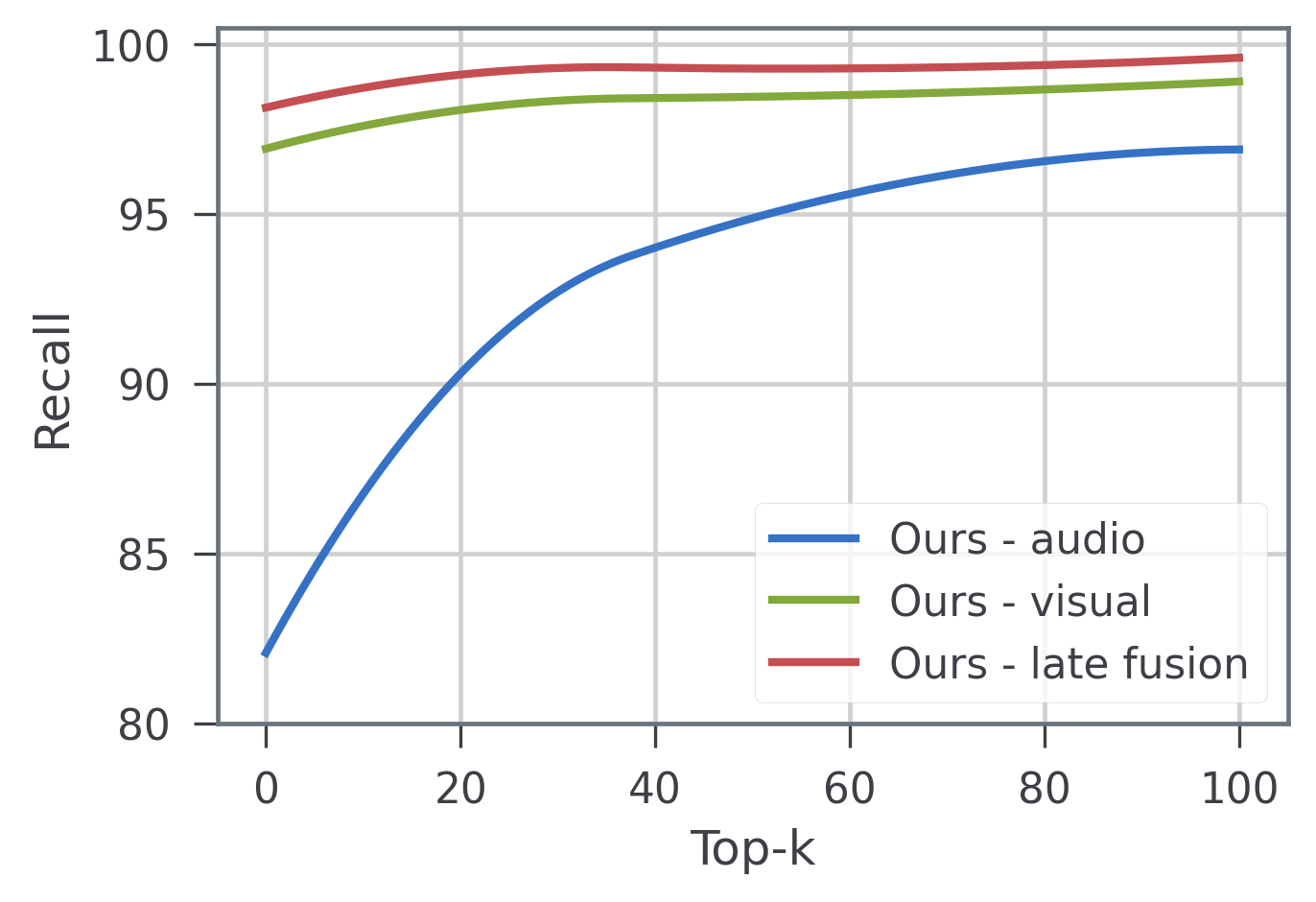}
    \caption{Recall curve in \% versus top-k of our audio, visual and late fusion models.}
    \label{fig:aud_recall}
\end{figure}

\subsection{Evaluating visual features}
We compare our method with two following baselines:
\begin{itemize}
    \item \textbf{ResNet50 (ImageNet)}\cite{he2016deep} - a generic model pretrained on ImageNet for image classification. We use the 2048-D output of the average pooling layer for our embedding. 
    \item \textbf{DeepComp}\cite{Black_2021_CVPR} - a Resnet50-based model that is trained to be robust against ImagenetC\cite{imagenetc} perturbation as well as photoshop manipulation.
\end{itemize}
While changing the backbone network, all other parameters, such as codebook size and search algorithm are fixed.
Table~\ref{tab:vis_main} shows the recall performance of our approach and the baselines on the VGGSound dataset. Although all 3 models have the same CNN architecture, ResNet50 (ImageNet) performs the worst since it is not exposed to the video transformations (sec.~\ref{sec:features_visual}) during training. DeepComp \cite{Black_2021_CVPR} has significantly better performance because it is trained to be robust against a number of perturbations on images. Finally, the proposed approach outperforms the rest at all recall levels. This demonstrates that robustness against particular perturbations could be achieved by exposing the model to the same augmentations during training. 

The partial matching performance on visual stream is depicted in Figure~\ref{fig:vis_qlen}. Overall, the retrieval is more challenging with shorter queries. The proposed method slightly underperforms DeepComp \cite{Black_2021_CVPR} at R@1 on very short queries, but outperforms on queries longer than 3 seconds (30\% length of the original videos) and always performs favorably at R@10 on all query lengths. In contrast, ResNet50 \cite{he2016deep} performs poorly again, with R@1 (R@10) ranges between 10-20\% (15-30\%) as the query length increases.

We report the robustness of our visual model against individual transformations in Figure~\ref{fig:vis_iaug}. Overall, there is not much difference between top-1 and top-100 recall, thanks to the re-ranking step (as demonstrated later in sec.~\ref{sec:abl}). Also, our model is most robust against common degradation (noise, encoding quality, pixelization) and minor content editing (text/shape overlay) with above 90\% recall at all levels, while being more sensitive to padding, rotation and color jitter.

\subsection{Evaluating audio features}
We compare our method with 3 following baselines:
\begin{itemize}
    \item \textbf{VGGish} \cite{vggish} - a VGG-based model trained on 70M audio clips to classify 3k audio categories. This model produces a 128-D descriptor per 0.96ms AW. 
    \item \textbf{AVENet} \cite{vggsound} - a ResNet18-based model released with the VGGSound dataset. This model produces a 512-D descriptor per 10-second audio.
    \item \textbf{AVENet (finetuned)} is the AVENet model finetuned using our data augmentations and loss. We keep the same model architecture and data processing but adopt our training procedure for fair comparison. 
\end{itemize}

\begin{table}
    \centering
    \begin{tabular}{l|ccc}
        Query set & R@1 & R@10 & R@100 \\ \hline
        Real & 0.918 & 0.972 & 0.999\\
        Fake & 0.912 & 0.966 & 0.998
    \end{tabular}
    \caption{DFDC search performance}
    \label{tab:dfdc}
\end{table}

The recall performance is reported in Table~\ref{tab:aud_main}. Our method outperforms the rest with a significant margin (21-25\% consistent improvement from the nearest competitor at all recall levels). Our finetuned version of AVENet also has better performance than other pretrained models, while VGGish \cite{vggish} performs the worst. 

The similar trend is observed in our experiment with query length (Figure~\ref{fig:aud_qlen}). Our method leads a 20\% gap ahead of AVENet and AVENet (finetuned) at all query lengths, except at 1-second queries where its R@1 score levels with AVENet.

Figure~\ref{fig:aud_iaug} shows the effects of individual audio perturbations on the recall performance of the proposed method. Our model is most sensitive to background noise while being less affected by other sources (c.f. Figure~\ref{fig:aud_aug}).


\begin{figure*}[t]
    \centering
    \includegraphics[width=1.0\linewidth,height=7cm]{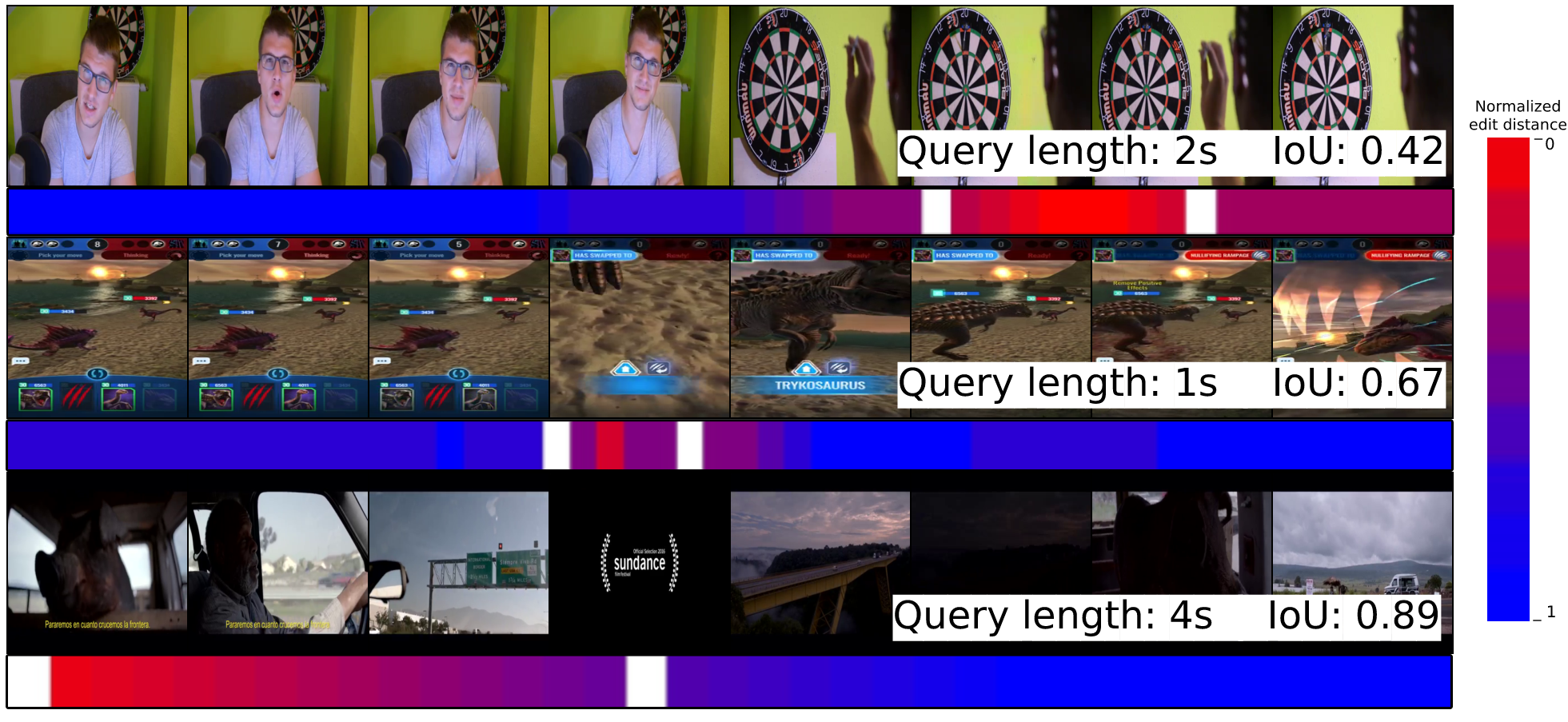}
    \caption{Localization of query clips within the top-1 retrieved video. For each row, the query clip is the video segment within the white bars of the original video (also subjected to random augmentations). In all 3 cases the original video is returned at top-1. The heatmap bar shows the edit distance between the query sequence of codewords and a same-length segment of the candidate video in sliding window fashion, which could be used to represent the confidence in localization of the query within the candidate video. IoU score can be computed over a thresholded heatmap and the ground truth location.}
    \label{fig:localisation}
\end{figure*}

\subsection{Audio-visual fusion}
We evaluate 3 fusion methods described in sec.~\ref{sec:fusion}. Table~\ref{tab:fusion} indicates the superiority of the late fusion method, with 98\% recall at top-1 and near perfect performance at top-100. Learned fusion method has better score than early fusion but lower than late fusion, probably because the unified embedding and inverted index are effectively twice as compact as the combined audio and visual. Another advantage of late fusion is that it enables querying of individual modality, for example in case an user only has single-modal data or prefer to retrieval an individual stream. Figure~\ref{fig:aud_recall} demonstrates that audio and visual retrieval has complementary effects, as the late fusion method improves performance versus any single-stream retrieval method.




\subsection{Video attribution on DFDC-Preview}
\label{sub:dfdc}
We employ our VGGSound-trained model together with the best fusion method (late fusion) and  test directly on the DFDC-Preview dataset. Despite the domain shift, our method performs extremely well on the public augmented-ready {\em real} set and the manipulated {\em fake} set whose transformations are unseen during training, as shown in Table~\ref{tab:dfdc}. On both sets we achieve above 91\% R@1 and the performance is on par with the VGGSound results at R@100 (c.f. Table~\ref{tab:fusion} late fusion), demonstrating the generalization of our approach.

For the {\em fake} set, we further visualize the manipulated area using the ICN method proposed in \cite{Black_2021_CVPR}. Figure~\ref{fig:dfdc_comp} shows the visualization between an example query and the top-1 retrieval video. This is useful in practice for a user to be able to attribute a query video back to its original source, localize its position in the source and highlight any possibly manipulated regions. 

\subsection{Ablation study}
\label{sec:abl}
We test the efficacy of our retrieval pipeline when stripping off one or several components (Table~\ref{tab:abl}). We first turn off the re-ranking stage (sec.~\ref{sec:lev}) and rank the results using only the TF-IDF relevance score. Without re-ranking the recall score significantly drops by 18\% at R@1. This indicates that re-ranking can promote the relevant video to the top of the ranking by leveraging the temporal sequence of codewords within a chunk (and sequence of chunks within a long video). Next, we further turn off TF-IDF and use only the histogram count of codewords in the inverted index to rank the videos. The performance further reduces by 3\% at R@1. 

\subsection{Localization of video segments}
We demonstrate an example of partial matching and localization of a video segment in Fig.~\ref{fig:localisation}. Since each video can be represented as a variable-length sequence of chunks and each chunk is a sequence of codewords, we can compute the similarity between the query sequence with a sequence of a candidate video using edit distance, thus be able to localize position of the query in the candidate video (at chunk-level for long videos and codeword-level for short videos). 

\begin{table}
    \centering
    \begin{tabular}{l|ccc}
        Method & R@1 & R@10 & R@100 \\ \hline
        proposed & 0.982 & 0.991 & 0.996 \\
        w/o re-ranking & 0.798 & 0.984 & 0.995 \\
        w/o TF-IDF + re-ranking & 0.764 & 0.982 & 0.992 
    \end{tabular}
    \caption{Ablation study when re-ranking with edit distance and TF-IDF is turned off.}
    \label{tab:abl}
\end{table}

\section{Conclusion}

We present a robust video provenance system capable of matching full or partial videos  circulating online without provenance metadata, to an authoritative copy of that video held within a trusted database.   Our matching process is robust to transformations of the video such as  changes in format, resolution, size, padding that video often undergoes during online redistribution.  We employ self-supervised contrastive learning to learn audio and visual features robust to such transformations, and match these using a vector quantization and inverse index.  We explored the relative benefits of early versus late fusion for accommodating both modalities, and proposed an edit-distance based re-ranking process for shortlisted results. For feature learning we showed that recently proposed augmentation sets for image similarity extend well to video for the authenticity use, and that recent techniques for visualizing manipulated regions in images may be effectively applied to video on a per-frame basis.

Future work could explore integration of our feature embeddings with emerging standards \cite{c2pa,cai,origin} and decentralized frameworks such as blockchain to enable decentralized content registries for tracing videos distributed online.  For example, in the draft technical specification of the cross-industry C2PA body \cite{c2pa} it is recognised that metadata will frequently be stripped from assets and that a `soft binding' (such as the proposed video matching solution) may be used re-unite assets with their originals in a trusted `provenance datastore'.  Since this implies a centralization of trust, it may be interesting to store this database in a decentralized fashion.  Such approaches are already being explored for example in the domain of archives and memory institutions \cite{archangel,bui2020archangel}.

\section*{Acknowledgement}

We thank Andy Parsons, Leonard Rosenthol and the Adobe Content Authenticity Initiative (CAI) for discussions.  This work was supported by Adobe Research via a PhD studentship and via the DECaDE Centre by EPSRC Grant Ref EP/T022485/1.

\bibliographystyle{ACM-Reference-Format}
\bibliography{bib}

\end{document}